\documentclass{article}

    \usepackage[final]{neurips_2021}

\usepackage[utf8]{inputenc} 
\usepackage[T1]{fontenc}    
\usepackage{hyperref}       
\usepackage{url}            
\usepackage{booktabs}       
\usepackage{amsfonts}       
\usepackage{nicefrac}       
\usepackage{microtype}      
\usepackage{xcolor}         

\usepackage{float}
\usepackage{wrapfig}

\usepackage{graphicx}
\usepackage{subfigure}

\title{Gamifying Math Education using Object Detection}

\author{%
  Yueqiu Sun \\
  Tangible Play\\
  Palo Alto, CA 94306 \\
  \texttt{yueqiu@tangibleplay.com} \\
   \And
   Rohitkrishna Nambiar \\
   Tangible Play \\
   Palo Alto, CA 94306 \\
   \texttt{rohit@tangibleplay.com} \\
   \And
   Vivek Vidyasagaran \\
   Tangible Play \\
   Palo Alto, CA 94306 \\
   \texttt{vivek@tangibleplay.com} \\
}

\begin{document}

\maketitle

\begin{abstract}

Manipulatives used in the right way help improve mathematical concepts leading to better learning outcomes. In this paper, we present a phygital (physical + digital) curriculum inspired teaching system for kids aged 5-8 to learn geometry using shape tile manipulatives. Combining smaller shapes to form larger ones is an important skill kids learn early on which requires shape tiles to be placed close to each other in the play area. This introduces a challenge of oriented object detection for densely packed objects with arbitrary orientations. Leveraging simulated data for neural network training and light-weight mobile architectures, we enable our system to understand user interactions and provide real-time audiovisual feedback. Experimental results show that our network runs real-time with high precision/recall on consumer devices, thereby providing a consistent and enjoyable learning experience.

\end{abstract}

\section{Introduction}

Manipulatives are objects used to represent abstract concepts explicitly and concretely [12, 19, 20, 21, 22]. They have been used to teach math at all grade levels for a while now [3]. In a comparative study [7] with first grade students, we see that the use of tangible manipulatives increases action possibilities which led to a positive impact in children's mathematical abilities. Using manipulatives not only improves mathematical concepts but also helps students construct their own mathematical understanding thereby providing a richer learning experience [5]. However, manipulatives on their own have very little meaning and provide maximum benefit when paired with teacher guidance in a comprehensive and well planned setting [18]. This is also not scalable outside of a school environment.

AR based solutions such as \textit{ARMath} [6] present a mobile AR platform that recognizes everyday objects and turns them into manipulatives. Mixed reality learning systems [8, 9, 11] combine manipulatives with virtual feedback. However, issues such as difficult hand-eye coordination and device stability limit the practical use of AR based methods. By combining physical manipulatives with a digital game using reflective artificial intelligence technology (RAIT) [1], we present a hands-on play-based learning system for teaching mathematics to kids. Our setup seen in Fig.~\ref{fig:setup} consists of a user tablet mounted on a base, an angled mirror called reflector, and shape tile manipulatives. RAIT through the reflector enables the front camera of the device to \textit{"see"} the table surface as the user interacts with the manipulatives. 

In this work, we focus on teaching geometry to grades 1-2 through a digital game using shape tile (e.g., semicircle, triangle, square) manipulatives. The core objective of the game is to create food ingredients from geometric shapes for digital characters. Some of the geometric skills that we cover are to identify and name shapes, to compose and partition shapes and to reason with shapes and their attributes. As an example, shape composition for a mushroom ingredient can be seen in Fig.~\ref{fig:manipulatives}. To provide feedback through the game, we need to detect the 2D pose of the shape tiles placed close to each other. Further, to scale to users across the globe, the system has to run real-time seamlessly on user devices from inexpensive low-end tablets to high-end ones. Our contributions are three part. First, we present a phygital system for learning geometry using shape tile manipulatives for grades 1-2. Second, we introduce a synthetic data generation system for neural network training and third, our proposed oriented object detection network addresses the aforementioned challenges with high precision/recall thereby providing an engaging, consistent and enjoyable learning experience.

\section{Setup}

\begin{figure}
\parbox{0.48\linewidth}{
    \centering
    \includegraphics[height=2in, width=0.45\columnwidth]{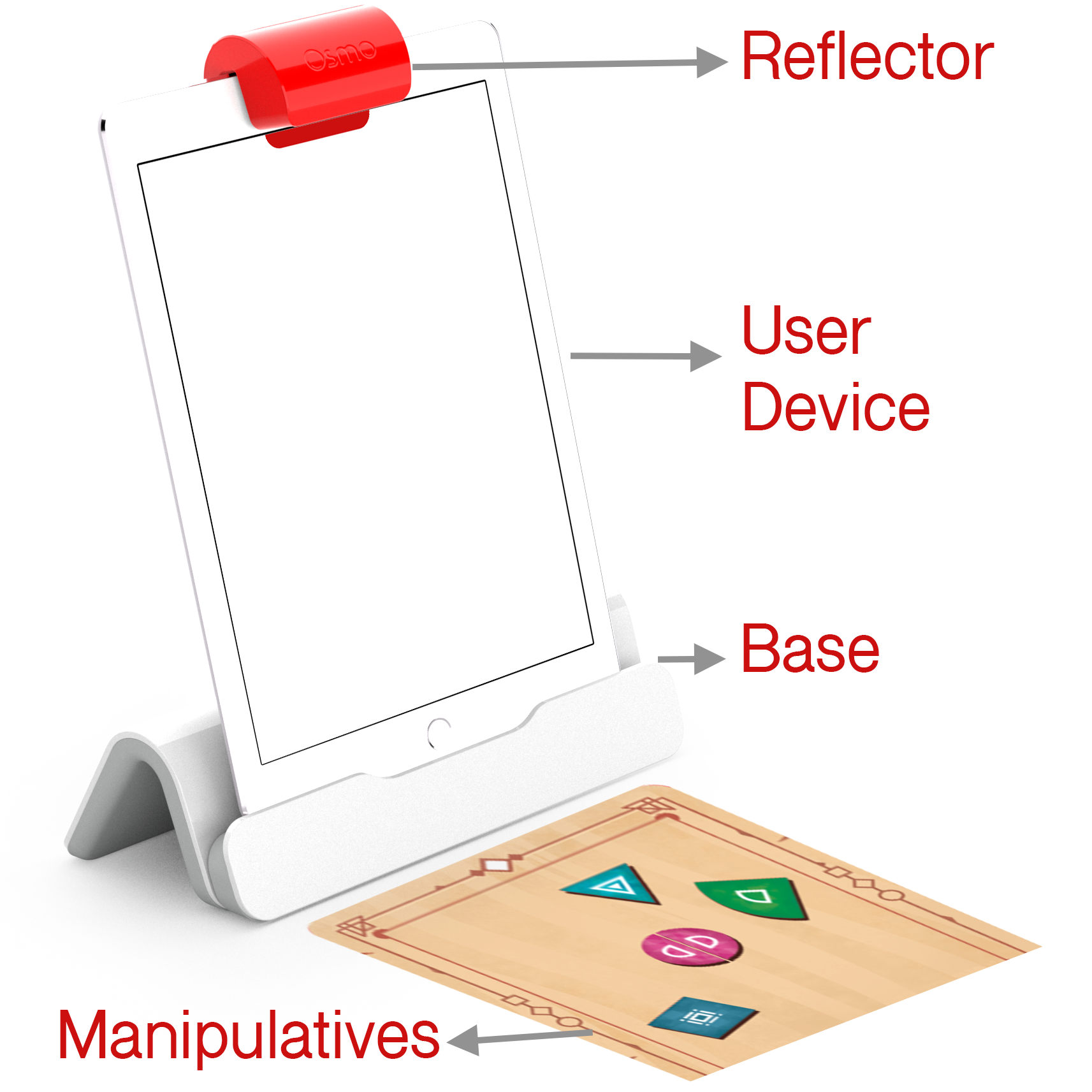}
    \caption{Phygital setup with a user device, a reflector and base. The forward facing camera is able to see the manipulatives placed on the tabletop due to the angled mirror inside the reflector.}
    \label{fig:setup}
}\hfill
\parbox{0.48\linewidth}{
    \centering
    \includegraphics[height=2in, width=0.30\columnwidth]{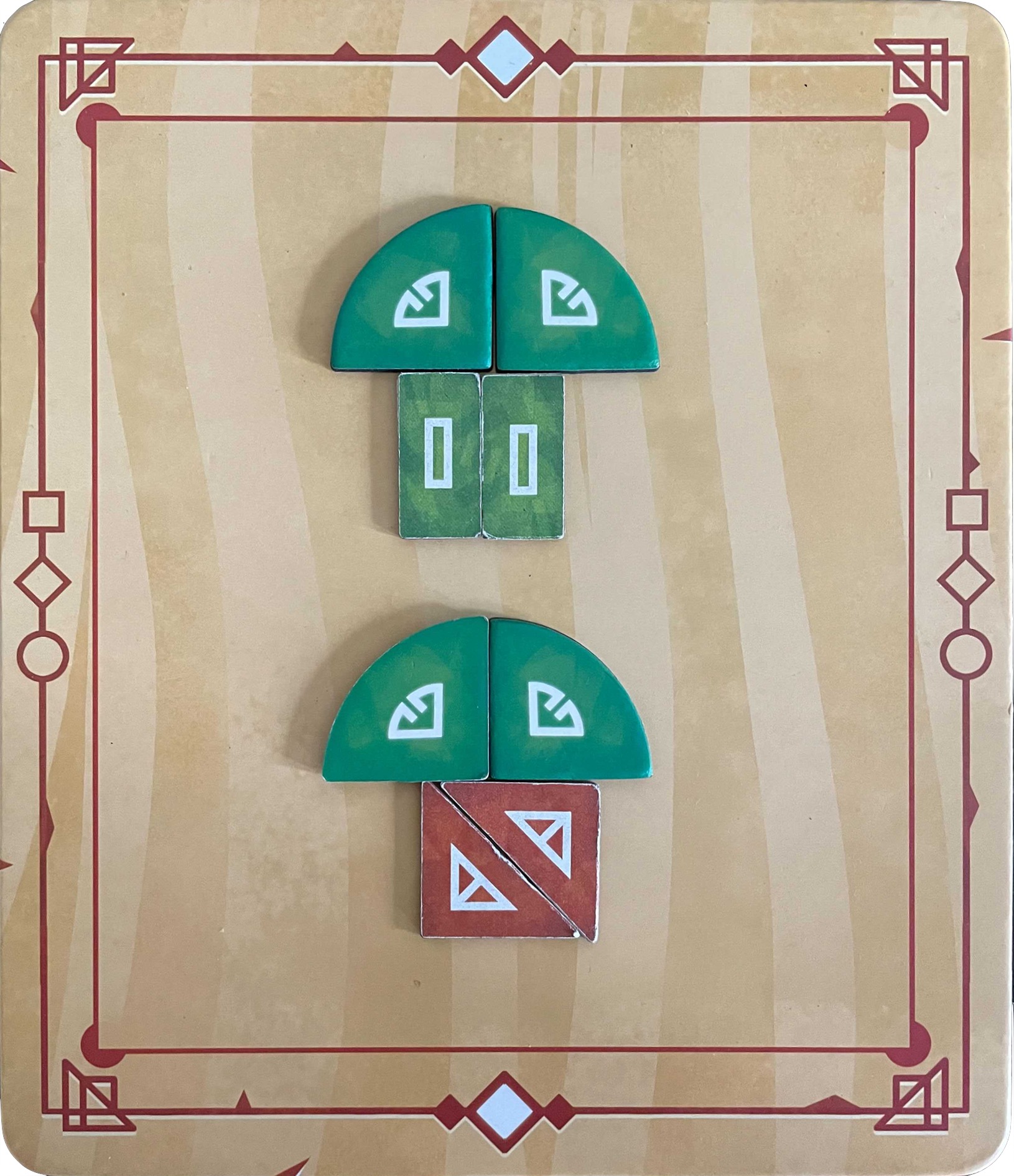}
    \caption{Shape tile manipulatives placed on a playmat. Smaller tiles are combined to form larger shapes. ex. two dark-green quarter circles and two red right angle triangles can be placed together to form a mushroom ingredient. }
    \label{fig:manipulatives}
}
\end{figure}

The hardware setup consists of a user device, a base, an angled mirror called reflector, and shape tile manipulatives. The device is mounted on the base and the reflector is placed on top of the front facing camera enabling the device to capture the tabletop play area as depicted in Fig.~\ref{fig:setup}. The manipulatives themselves are designed to be simple and relatively inexpensive. They are made from cardboard and do not have any electronic or magnetic components. Their appearance is designed to match the theme of the game thereby avoiding any obvious visual markers or codes which provides a cohesive user experience. We use the terms tiles and manipulatives interchangeably throughout the paper.

\section{Method}

Apart from the physical manipulatives, our system consists of a digital game and a computer vision module. A typical user flow begins with the digital game character asking the user to build an ingredient using geometric shape tiles. Kids interact with the system by placing and moving the tiles on the play area. Our unique hardware setup enables us to compute an accurate bird's-eye view (BEV) image used in our CV pipeline. The game through the CV module predicts the \textit{id}, \textit{location}, and \textit{orientation} of the manipulatives and provides audiovisual feedback to the user based on certain rules. For example, in Fig. ~\ref{fig:manipulatives}, the stem of the mushroom ingredient can be built using two green rectangle tiles or two red right angled triangles. In the following, we will elaborate on the data problem and the architecture used for on-device object detection with orientation prediction.

\subsection{Data Scarcity}

\begin{figure}
    \includegraphics[width=.24\textwidth]{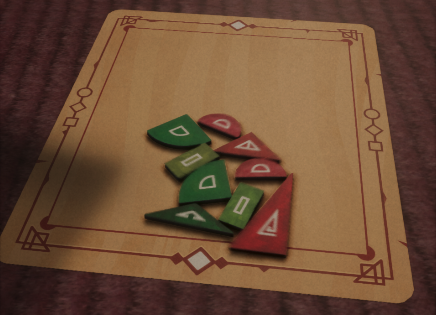}\hfill
    \includegraphics[width=.24\textwidth]{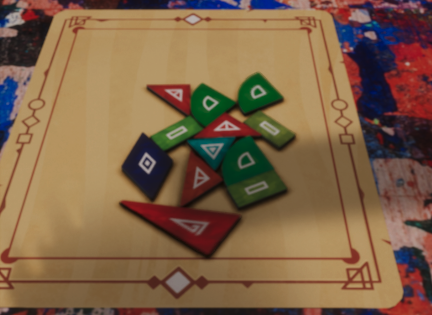}\hfill
    \includegraphics[width=.24\textwidth]{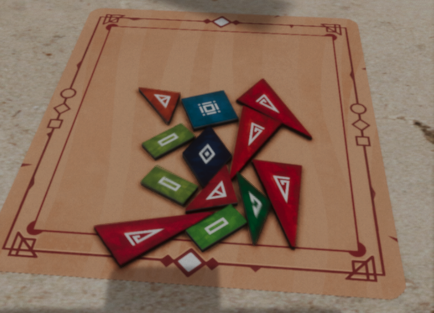}\hfill
    \includegraphics[width=.24\textwidth]{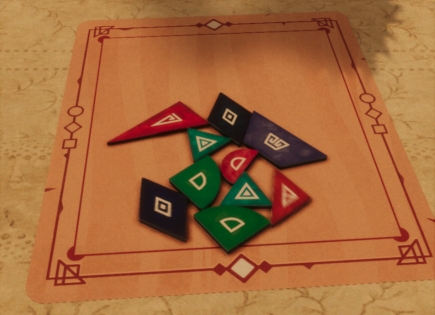}\hfill
    \caption{Simulated data generation using Unity Perception SDK [24] pipeline. We primarily vary tile and scene configurations.}\label{fig:simulateddata}
\end{figure}

Our product is fully COPPA compliant, and that means we don't collect images from our customers. However, we do have the ability to collect limited data from beta testing and by using the system ourselves, but this in most cases isn't enough for our network training efforts. To tackle this challenge, we use data augmentation along with synthetic data generated using 3D models of our manipulatives in a Unity game engine [23]. We use a custom built synthetic data generation system that takes images of the manipulatives, generates 3D meshes as they would look in real life, and places them in different configurations on a virtual tabletop. We vary each scene using different camera parameters and lighting conditions as shown in Fig.~\ref{fig:simulateddata}. This enables large amounts of annotated data to be generated for pre-training making the model robust to difficult conditions in the real world.

\subsection{Network Architecture}

\begin{figure}[h]
  \centering
  \includegraphics[width=1\textwidth]{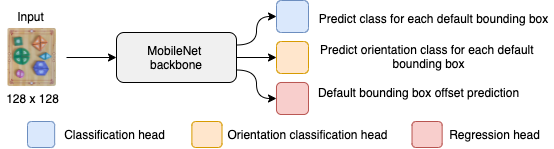}
  \caption{Network Architecture}
  \label{fig:network1}
\end{figure}

To provide a smooth and prompt user experience, we need to run network inference on-device in real-time. This constraint requires us to build lightweight neural networks which are time and memory efficient. Our network is based on the SSD [15] object detection architecture with MobileNet [13] backbone as shown in Fig. \ref{fig:network1}. The outputs from the MobileNet backbone are fed to two classification heads and one regression head to predict object class, orientation class and the bounding box offsets. We will discuss more details in the following sections and in the appendix.

\paragraph{Classification head:}
The classification head takes the output produced by the backbone network at different scales and outputs the classification score for each object class. The confidence is obtained using a softmax operation and the loss is computed using focal loss [16].

\paragraph{Orientation head:}
Composition of shapes ie. using smaller shape tiles to form a larger shape is a key skill we focus on in this game. For this, apart from detecting the location of the tiles, we also need to detect orientation with high accuracy. To predict continuous orientation of objects in the range $0 - 360^{\circ}$, we cannot directly perform regression because of its discontinuity in the representation space. We discretize the $0 - 360^{\circ}$ range into $N$ unique orientations which are $G = 360 / N$ degrees apart and convert the continuous prediction task into an N-class classiﬁcation task [14]. Each training sample is assigned one of the $N$ class labels based on its proximity to the discretized orientation bin. We set $N = 48$, where the orientation head outputs a confidence score for each class. The loss is calculated using focal loss. A potential downside of this approach is the loss of information introduced by the discretization step. However, we find this approach more stable in practice compared to a regression based approach. 

\paragraph{Regression head:}
The regression head also takes the output produced by MobileNet backbone at difference scales and outputs the prediction regression offsets for each bounding box.

\section{Experimentation and Results}

We evaluate the performance of our models on both the F-score and inference time as both are crucial for a smooth user experience. For the backbone network, we experiment with MobileNet and VGG architecture. Within models of the same backbone, version 1 is wider compared to version 2 (see architecture in Appendix A). For training, we use negative sampling ratio as 1 and set the optimizer to ADAM with learning rate as 0.0003. The models are pre-trained on 15,000 simulation images and fine-tuned on 5800 real world images. Roughly 20\% of the real images from Beta testing volunteers are used for hyperparameter tuning. 

\begin{table}[h]
  \caption{Detection results using difference models}
  \label{model-results-table}
  \centering
  \begin{tabular}{lllll}
    \toprule
    \cmidrule(r){1-2}
    Model     & Recall & Precision & F-score & Inference time(ms)\\
    \midrule
    Mobilenet Backbone 1&  99.04 &99.57 & 99.30 & 22.25  \\
    Mobilenet Backbone 2 &  99.57&98.83&99.20 & 10.81  \\
    VGG Backbone 1 &   98.83   & 99.46 & 99.14 & 66.86\\
    VGG Backbone 2     &    99.36 & 97.99 & 98.67 & 17.83 \\
    \bottomrule
  \end{tabular}
\end{table}

\begin{figure}[h]
    \centering
    \subfigure[]{\includegraphics[width=.24\textwidth]{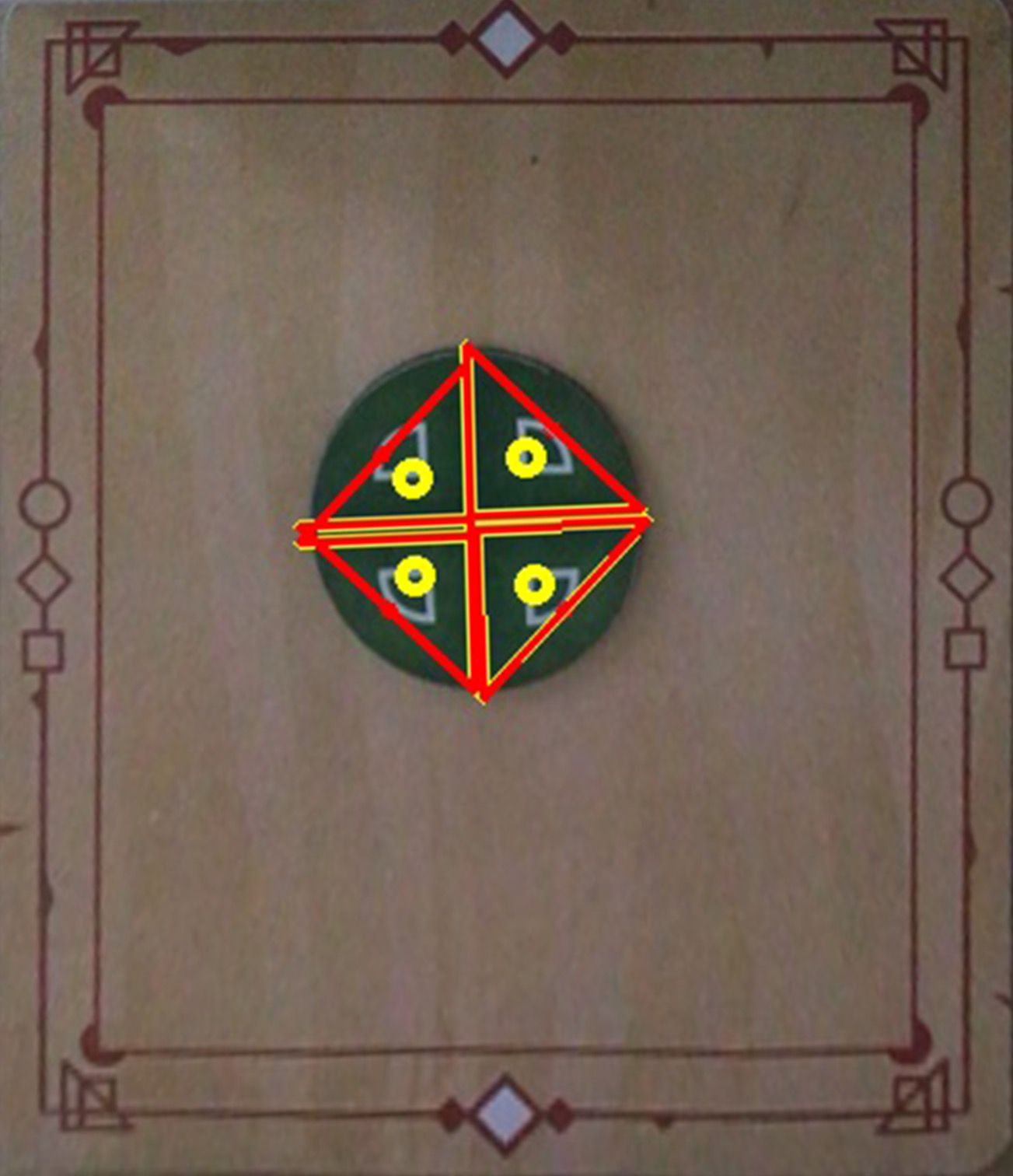}}
    \subfigure[]{\includegraphics[width=.24\textwidth]{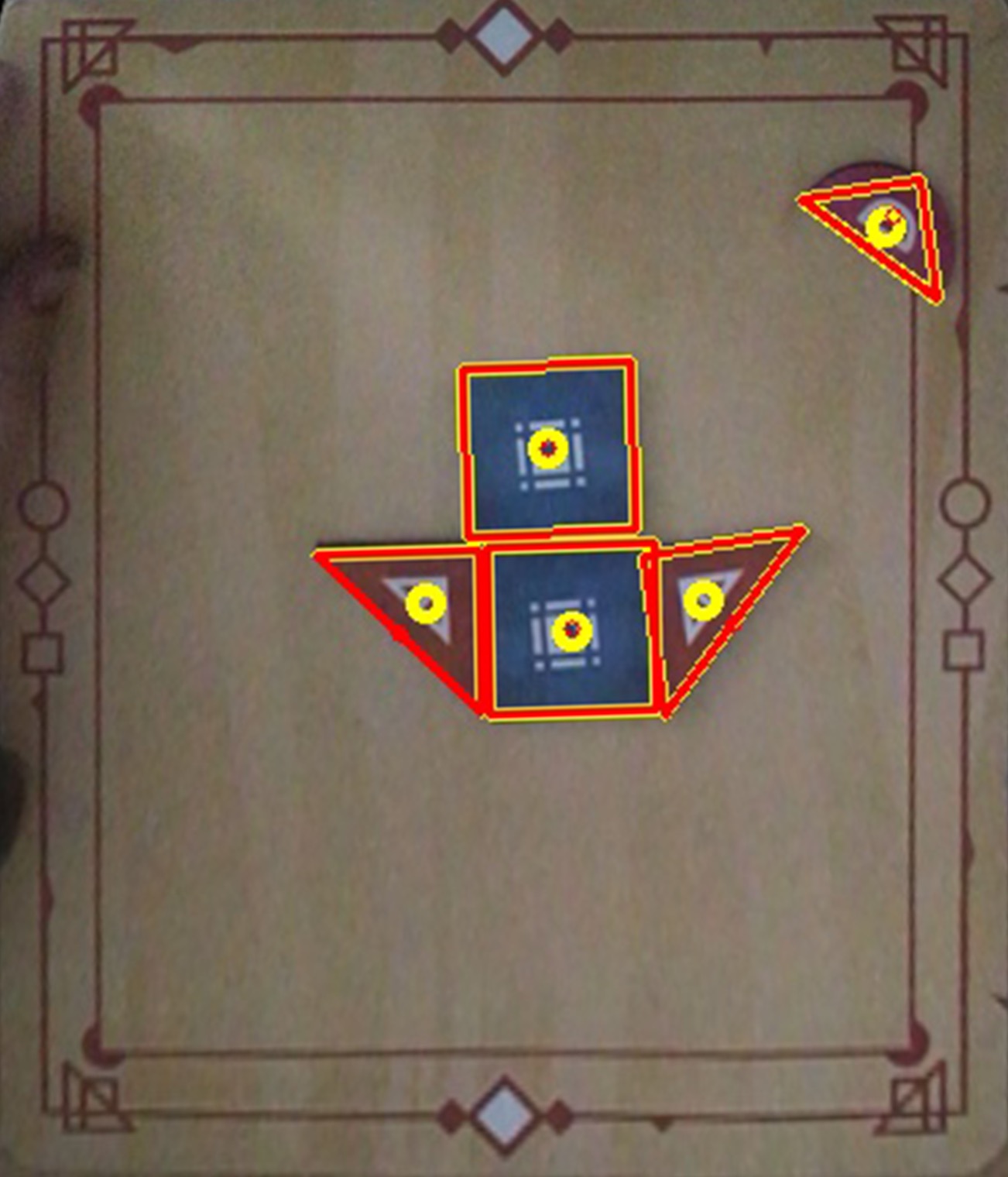}}
    \subfigure[]{\includegraphics[width=.24\textwidth]{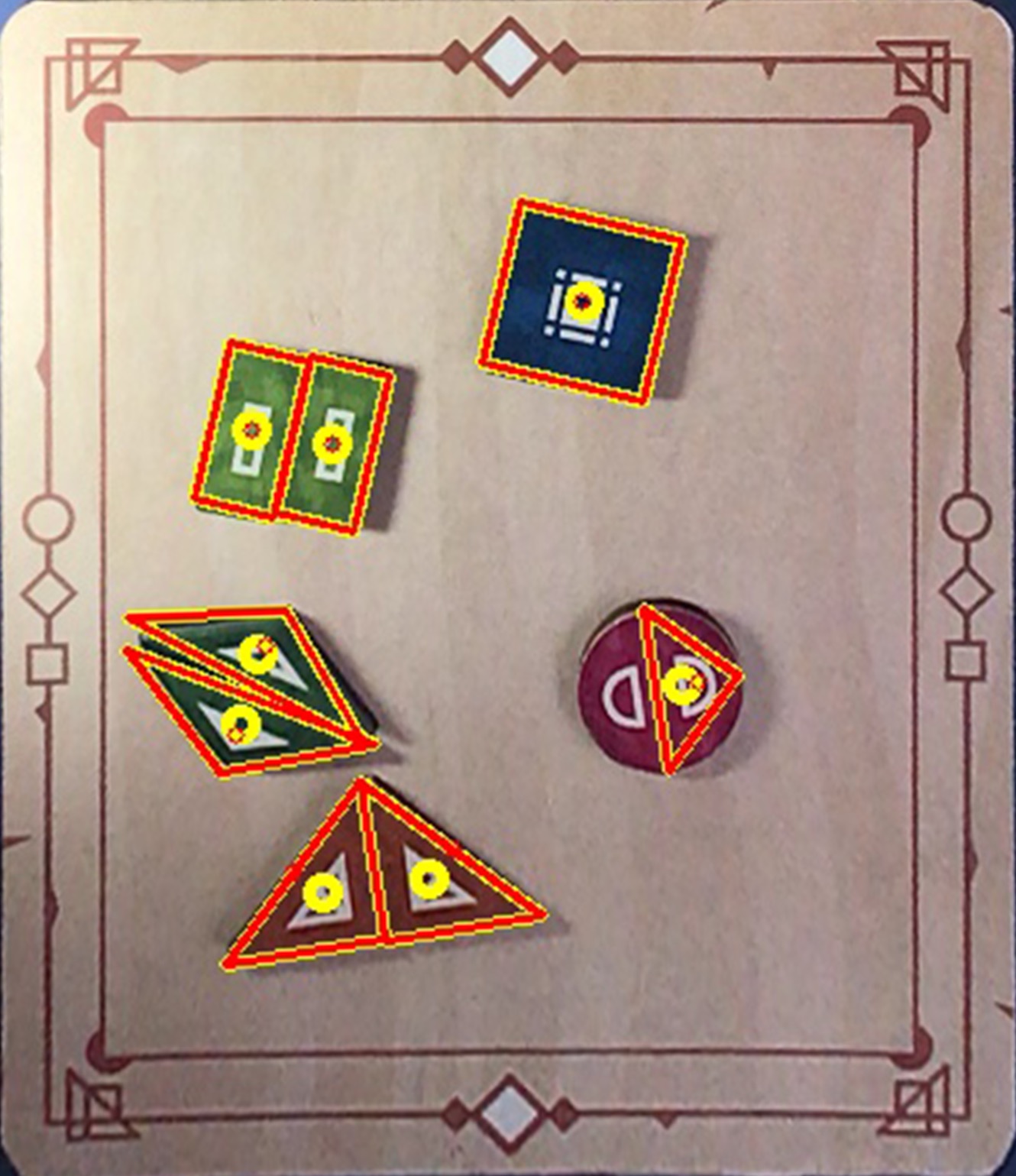}}
    \subfigure[]{\includegraphics[width=.24\textwidth]{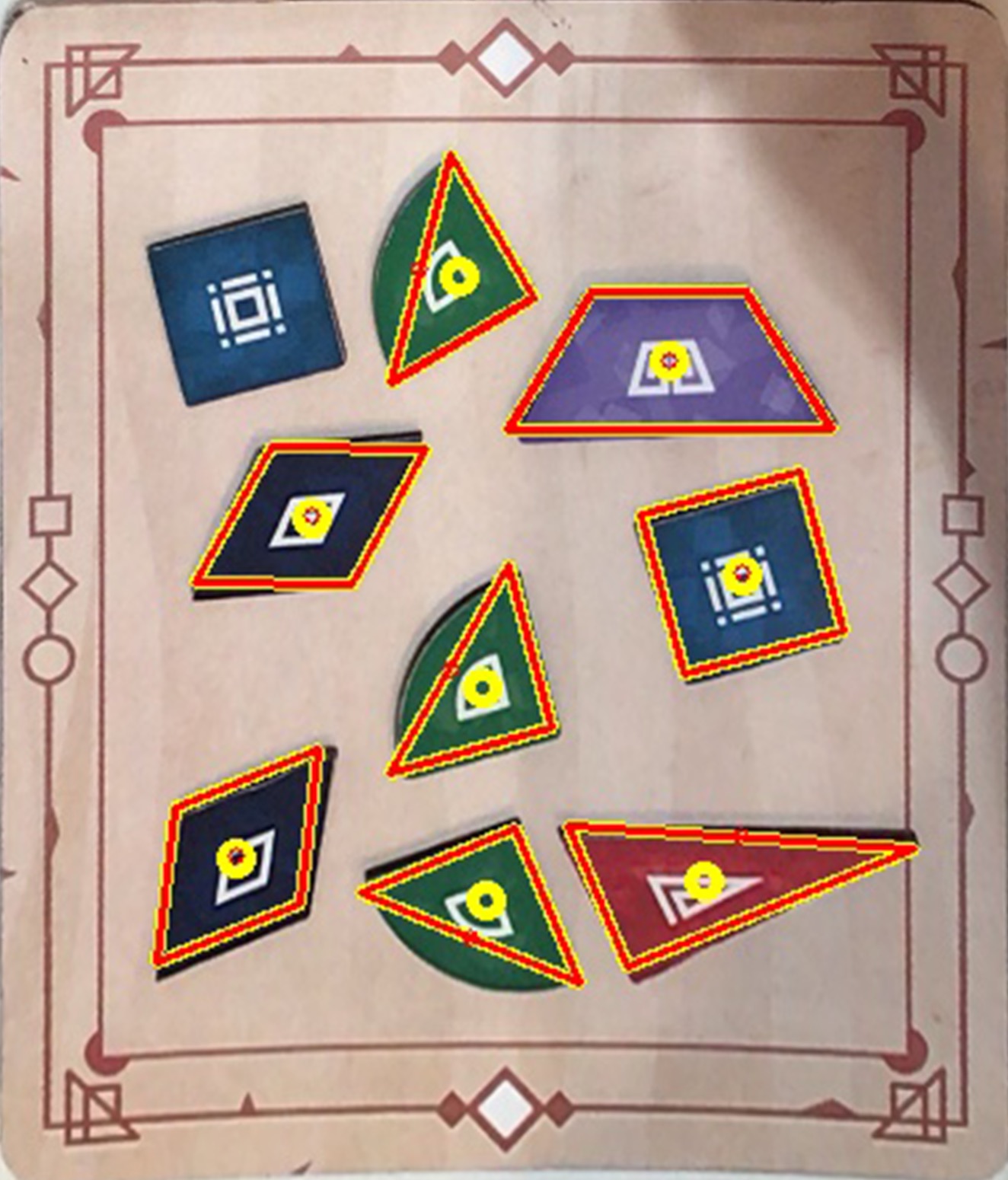}}
    \caption{Model predictions mapped to shape polygons on test set. In images (a) and (b) we see that all the tiles are correctly detected whereas (c) and (d) have missed detections.}
    \label{fig:test_results}
\end{figure}

During forward pass, the network predictions are passed through decode layers that consists of non-maximum suppression and an output layer that maps network outputs to individual vertices of the shape polygons. The precision/recall and F-score metrics are obtained using the test set consisting of images captured from play sessions with different users in different environment conditions. The predicted vertices are matched to the ground-truth vertices to obtain the above metrics. From Table \ref{model-results-table}, we see that models with MobileNet backbone achieve a better F-score compared to the models with VGG backbone while running with a much shorter inference time. We show model predictions mapped to shape polygons drawn on a few images from the test set in Fig. ~\ref{fig:test_results}.

\section{Conclusion}

By bridging the gap between a digital game and tangible interactions with shape tile manipulatives through reflective AI, we introduce a novel way of interaction for kids to engage and learn geometry while having fun. We presented a light-weight neural network for dense oriented object detection that supports real-time performance on a wide range of consumer tablets. Using synthetic data, we generate random shape compositions that are difficult to obtain in real images making the network more robust to real-world variations. By adopting a game-centric approach, we keep the experience non-repetitive and engaging for kids. For future work, we would like to cover different skills such as addition, subtraction, measurement and counting, etc.

\begin{ack}
We thank everyone at Osmo/Tangible Play that was involved with the design and development of the hardware manipulatives and digital game as well as the user research studies. We thank Anoop Rajagopal, Nrupatunga and Heidy Maldonado for their initial feedback.
\end{ack}

\section*{References}

\medskip

{
\small

[1] Osmo\ --\ Award-Winning Educational Games System for iPad.\ https://www.playosmo.com/\  (Accessed on 09/20/2021)

[2] D'angelo, F., \& Iliev, N. (2012). Teaching Mathematics to Young Children through the Use of Concrete and Virtual Manipulatives. {\it Online Submission}.

[3] Boggan, M., Harper, S., \& Whitmire, A. (2010). Using Manipulatives to Teach Elementary Mathematics. {\it Journal of Instructional Pedagogies}, 3.

[4] Kosko, K. W., \& Wilkins, J. L. (2010). Mathematical communication and its relation to the frequency of manipulative use. {\it International Electronic Journal of Mathematics Education}, 5(2), 79-90.

[5] Cockett, A., \& Kilgour, P. W. (2015). Mathematical manipulatives: Creating an environment for understanding, efficiency, engagement, and enjoyment. {\it Teach Collection of Christian Education}, 1(1), 5.

[6] Kang, S., Shokeen, E., Byrne, V. L., Norooz, L., Bonsignore, E., Williams-Pierce, C., \& Froehlich, J. E. (2020, April). ARMath: augmenting everyday life with math learning. In {\it Proceedings of the 2020 CHI Conference on Human Factors in Computing Systems} (pp. 1-15).

[7] Pires, A. C., González Perilli, F., Bakała, E., Fleisher, B., Sansone, G., \& Marichal, S. (2019, September). Building blocks of mathematical learning: Virtual and tangible manipulatives lead to different strategies in number composition. In {\it Frontiers in Education} (Vol. 4, p. 81). Frontiers.

[8] Marichal, S., Rosales, A., Perilli, F. G., Pires, A. C., Bakala, E., Sansone, G., \& Blat, J. (2017, September). Ceta: designing mixed-reality tangible interaction to enhance mathematical learning. In {\it Proceedings of the 19th International Conference on Human-Computer Interaction with Mobile Devices and Services} (pp. 1-13).

[9] Almukadi, W., \& Stephane, A. L. (2015, November). BlackBlocks: tangible interactive system for children to learn 3-letter words and basic math. In {\it Proceedings of the 2015 International Conference on Interactive Tabletops \& Surfaces} (pp. 421-424).

[10] Scarlatos, L. L. (2006). Tangible math. {\it Interactive Technology and Smart Education}.

[11] Scarlatos, L. L., Dushkina, Y., \& Landy, S. (1999, May). TICLE: A tangible interface for collaborative learning environments. In {\it CHI'99 Extended Abstracts on Human Factors in Computing Systems} (pp. 260-261).

[12] Moyer, P. S. (2001). Are we having fun yet? How teachers use manipulatives to teach mathematics. {\it Educational Studies in mathematics}, 47(2), 175-197.

[13] Howard, A. G., Zhu, M., Chen, B., Kalenichenko, D., Wang, W., Weyand, T., ... \& Adam, H. (2017). Mobilenets: Efficient convolutional neural networks for mobile vision applications. {\it arXiv preprint arXiv:1704.04861}.

[14] Hara, K., Vemulapalli, R., \& Chellappa, R. (2017). Designing deep convolutional neural networks for continuous object orientation estimation. arXiv preprint {\it arXiv:1702.01499}.

[15] Liu, W., Anguelov, D., Erhan, D., Szegedy, C., Reed, S., Fu, C. Y., \& Berg, A. C. (2016, October). SSD: Single shot multibox detector. {\it In European conference on computer vision} (pp. 21-37). Springer, Cham.

[16] Lin, T. Y., Goyal, P., Girshick, R., He, K., \& Dollár, P. (2017). Focal loss for dense object detection. {\it In Proceedings of the IEEE international conference on computer vision} (pp. 2980-2988).

[17] Barron, J. T. (2019). A general and adaptive robust loss function. {\it In Proceedings of the IEEE/CVF Conference on Computer Vision and Pattern Recognition} (pp. 4331-4339).

[18] Laski, E. V., Jor’dan, J. R., Daoust, C., \& Murray, A. K. (2015). What makes mathematics manipulatives effective? Lessons from cognitive science and Montessori education. {\it SAGE Open}, 5(2), 2158244015589588.

[19] Zuckerman, O. (2010). Designing digital objects for learning: lessons from Froebel and Montessori. {\it International Journal of Arts and Technology}, 3(1), 124-135.

[20] Horn, M. S. (2018). Tangible interaction and cultural forms: Supporting learning in informal environments. {\it Journal of the Learning Sciences}, 27(4), 632-665.

[21] Horn, M. S., Crouser, R. J., \& Bers, M. U. (2012). Tangible interaction and learning: the case for a hybrid approach. {\it Personal and Ubiquitous Computing}, 16(4), 379-389.

[22] Brosterman, N. (1997). {\it Inventing Kindergarten} (1st ed.). Harry N. Abrams.

[23] Unity Game Engine - www.unity.com. (Accessed on 10/06/2021)

[24] Unity Perception SDK - \ https://github.com/Unity-Technologies/com.unity.perception\ 
}

\end{document}